\documentclass[english]{lni}
\newcommand{\metricTTL}{48\,\text{h}}
\newcommand{\metricDatasetSize}{385}
\newcommand{\metricPairCount}{73,920}
\newcommand{\metricMCRuns}{1,000}

\newcommand{\metricDimFull}{768}
\newcommand{\metricDimMRL}{128}
\newcommand{\metricSVDDim}{99}
\newcommand{\metricSVDVariance}{95\%}

\newcommand{\metricLatencyFull}{0.337 \pm 0.024}
\newcommand{\metricLatencyMRL}{0.069 \pm 0.019}

\newcommand{\metricSpeedupFactor}{4.85-fold}
\newcommand{\metricSpeedupTimes}{4.85\times}
\newcommand{\metricSavingsPercent}{79.4\%}
\newcommand{\metricSavingsOver}{79\%}

\newcommand{\metricMAE}{0.036 \pm 0.022}
\newcommand{\metricMAEVal}{0.036}
\newcommand{\metricPreservationPercent}{96.4\%}

\usepackage{booktabs}
\usepackage{graphicx}
\usepackage{amsmath}

\usepackage{amssymb}
\usepackage{mathtools}

\usepackage{amsthm}
\usepackage{tikz}
\usepackage{pgfplots}
\pgfplotsset{compat=1.18}
\usetikzlibrary{shapes.geometric, arrows.meta, positioning, calc, patterns}
\usepackage{algorithm}
\usepackage{algpseudocode}
\usepackage{float}
\usepackage{placeins}

\begin{document}

\title[Kairos: Numerically Robust News Recommendation]{Kairos: Numerically Robust News Recommendation under Item Cold-Start via Cholesky-based LinUCB}

\author[1]{Finn Hertsch}{Hertsch.Finn-24@stud.dhbw-ravensburg.de}{0009-0004-1692-8134}

\affil[1]{DHBW Ravensburg\\School of Business, Data Science and AI\\Marienplatz 2, 88212 Ravensburg\\Germany}

\maketitle

\begin{abstract}
  Algorithmic news personalization in regional markets often fails because modern deep learning models require massive interaction data while real-world news has a short Time-to-Live ($\text{TTL} < \metricTTL$) and shallow article pools. This structural item cold-start deprives collaborative filtering of the data needed for robust modeling.
  This paper presents \emph{Project Kairos}, a framework that bridges this data scarcity through a contextual online learning approach (LinUCB). To ensure numerical integrity for continuous operation, Kairos replaces error-prone Sherman-Morrison inversions with direct rank-1 updates of Cholesky factors. This preserves the positive definiteness of the covariance matrix even under ill-conditioned data scenarios. Simultaneously, Matryoshka Representation Learning (MRL) integration addresses inference latency.
  Empirical evaluations based on the Tagesschau API demonstrate that exploiting semantic redundancy in the feature space achieves a \metricSpeedupFactor{} efficiency gain without significantly compromising ranking precision. Kairos thus provides a blueprint for high-performance recommendation systems in resource- and data-constrained environments.
\end{abstract}


\section{Introduction}
Through the growing flood of digital news, it becomes increasingly challenging for users to maintain an overview. To avoid information overload, precise algorithmic filtering is indispensable today \cite{Wu23}. Modern recommendation systems aim to select a user-specific choice from a large number of timely published articles, considering both individual preferences and temporal relevance.

In the news domain, article catalog dynamics pose a fundamental challenge. Unlike static catalogs in e-commerce or video streaming, news articles display extremely short half-lives. Empirical studies on large-scale datasets such as MIND show that the majority of user interactions occur within less than \metricTTL{} post-publication \cite{RD22}. Afterwards, article utility decays rapidly, defining a restricted Time-to-Live (TTL) \cite{ZS25}.

\subsection{The Item Cold-Start Paradox}
This short lifecycle creates a fundamental discrepancy for classical collaborative filtering. Approaches like Neural Collaborative Filtering (NCF) \cite{He17} or matrix factorization rely on high interaction density to learn robust latent representations.

In dynamic news environments, a paradox emerges: the time required to gather sufficient interaction coverage for classical methods often exceeds the content's relevance window \cite{RD22}. In regional corpora with modest article throughput per \metricTTL{} window, interaction density per item rarely reaches the critical mass needed for neural approaches. This favors popularity bias, systematically ignoring the long tail \cite{ABM19}.

\subsection{Proposed Approach and Contributions}
The framework \emph{Project Kairos}\footnote{Source code and experiments: \url{https://github.com/F1nnSBK/Project-Kairos}} addresses this issue by transitioning from passive data-driven modeling to active online learning via contextual bandits \cite{Li10}. Rather than waiting for interaction density, the implemented Linear Upper Confidence Bound (LinUCB) algorithm uses contextual embeddings to estimate recommendation payoffs while modeling uncertainty (exploration).

This paper contributes:
\begin{enumerate}
  \item \textbf{Numerical Stability in High-Load Environments:} Demonstrating that conventional LinUCB implementations relying on Sherman-Morrison matrix inversion are prone to rounding errors. Kairos replaces this with direct Cholesky factor updates, preserving SPD structure implicitly.
  \item \textbf{Adaptive Inference Architecture:} Integrating Matryoshka Representation Learning (MRL) \cite{Ku22} to dynamically adjust inference compute load, enabling efficient candidate generation with minimal precision loss.
\end{enumerate}

\begin{figure}[!htbp]
  \centering
  \begin{tikzpicture}[
      node distance=0.45cm and 1.2cm,
      font=\sffamily\small,
      box/.style={draw=black!80, thick, fill=gray!5, minimum width=4.0cm, minimum height=0.55cm, align=center, rounded corners=2pt},
      sidebox/.style={draw=black!80, thick, fill=gray!12, minimum width=3.0cm, minimum height=0.55cm, align=center, rounded corners=2pt},
      arrow/.style={-Stealth, thick}
    ]
    \node [box] (api) {Tagesschau API};
    \node [box, below=of api] (prep) {Article Preprocessing};
    \node [box, below=of prep] (mrl) {MRL Embedding Model (\metricDimFull d)};
    \node [box, below=of mrl] (mips) {MIPS Retrieval (\metricDimMRL d Subspace)};
    \node [box, below=of mips] (topk) {Top-$k$ Candidates};
    \node [box, below=of topk] (linucb) {LinUCB Re-Ranking (Cholesky)};
    \node [sidebox, right=1.0cm of linucb] (reward) {Reward Signal\\(Clicks / Interaction)};
    \node [box, below=of linucb] (rec) {Personalized Recommendations};

    \draw [arrow] (api) -- (prep);
    \draw [arrow] (prep) -- (mrl);
    \draw [arrow] (mrl) -- (mips);
    \draw [arrow] (mips) -- (topk);
    \draw [arrow] (topk) -- (linucb);
    \draw [arrow] (linucb) -- (rec);
    \draw [arrow] (reward) -- (linucb);
  \end{tikzpicture}
  \caption{Kairos system architecture. Candidate generation runs in the reduced Matryoshka subspace ($m=\metricDimMRL$), while final ranking is performed by a Cholesky-based LinUCB learner with incremental online updates.}
  \label{fig:arch}
\end{figure}
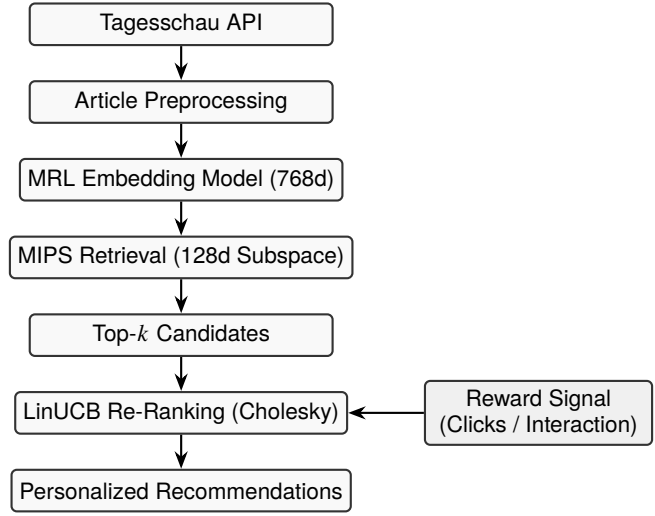

The overall framework architecture is outlined in \Cref{fig:arch}. Empirical evaluation data for numerical stability and inference performance is sourced from the Tagesschau API.

\section{Methodology}
Project Kairos pursues an integrative approach combining numerical stability in online optimization with highly efficient feature representations. The mathematical formulation of the exploration module and the architectural design of scalable feature extraction are derived below.

\subsection{LinUCB and Geometric Interpretation of Uncertainty}
The LinUCB algorithm assumes a linear relationship between contextual feature vectors $x_{t,a} \in \mathbb{R}^d$ of action $a$ at time $t$ and expected user rewards \cite{Li10}. Item selection maximizes the payoff term $p_{t,a}$, representing an upper confidence bound:
\begin{equation}
  p_{t,a} = \hat{\theta}_t^\top x_{t,a} + \alpha \sqrt{x_{t,a}^\top A_t^{-1} x_{t,a}}
  \label{eq:linucb}
\end{equation}
Here $\hat{\theta}_t \in \mathbb{R}^d$ denotes the estimated user preference vector, $\alpha > 0$ is the exploration hyperparameter, and $A_t \in \mathbb{R}^{d \times d}$ is the cumulative context covariance matrix. The covariance matrix accumulates over historical observations according to:
\begin{equation}
  A_t = \sum_{\tau=1}^{t-1} x_{\tau,a_\tau} x_{\tau,a_\tau}^\top + I_d
\end{equation}
where $I_d$ is the $d$-dimensional identity matrix for ridge regularization. The outer product $x_{\tau,a} x_{\tau,a}^\top \in \mathbb{R}^{d \times d}$ forms the outer product of the feature vector with itself. Since $x x^\top$ is symmetric ($(x x^\top)^\top = x x^\top$), $A_t = A_t^\top$ holds, forming a symmetric positive definite (SPD) covariance matrix. The inverse $A_t^{-1}$ corresponds to the precision matrix.

The quadratic term $x_{t,a}^\top A_t^{-1} x_{t,a}$ inside the square root in \Cref{eq:linucb} can be interpreted mathematically as the weighted Mahalanobis distance of context vector $x_{t,a}$ in the space defined by precision matrix $A_t^{-1}$. Geometrically, it describes a confidence ellipsoid around parameter estimates in $d$-dimensional space \cite{LS20}.

\begin{figure}[htbp]
  \centering
  \begin{tikzpicture}[scale=1.15, font=\sffamily\normalsize]
    \draw[thin, pattern=north east lines, pattern color=gray!35,
    rotate around={20:(2.8,2.2)}] (2.8,2.2) ellipse (2.2cm and 1.3cm);
    \draw[thin, fill=white, rotate around={20:(2.8,2.2)}] (2.8,2.2) ellipse (0.9cm and 0.5cm);

    \foreach \point in {(2.6,2.1),(2.9,2.3),(3.0,2.0),(2.7,2.4),(2.8,2.1),(2.5,2.2),(2.9,2.2),(2.7,2.0),(3.1,2.1),(2.8,2.3)} {
      \fill[black] \point circle (1.3pt);
    }

    \draw[-Stealth, thick] (2.8,2.2) -- (4.6,3.1) node[midway, above left, font=\small] {$x_{t,a}$};
    \fill[black] (4.6,3.1) circle (1.8pt);
    \fill[black] (2.8,2.2) circle (1.8pt);
    \node[below left, font=\small] at (2.8,2.2) {$\hat{\theta}_t$};

    \node[align=left, font=\small] (hd) at (0.6,4.3) {high data density};
    \draw[thin, -Stealth] (hd.south east) -- (2.4,2.5);

    \node[align=left, font=\small] (eb) at (5.7,3.9) {high\\exploration bonus};
    \draw[thin, -Stealth, shorten >= 2.5pt] (eb.200) -- (4.6,3.1);

    \node[align=left, font=\small] (gd) at (5.4,0.7) {low\\data coverage};
    \draw[thin, -Stealth] (gd.north west) -- (4.2,1.5);

    \draw[-Stealth, thick] (0.0,-0.2) -- (0.0,1.1) node[above] {$x_2$};
    \draw[-Stealth, thick] (0.0,-0.2) -- (1.2,-0.2) node[right] {$x_1$};
    \node[below left, font=\small] at (0.0,-0.2) {$0$};
  \end{tikzpicture}
  \caption{Geometric interpretation of the LinUCB exploration term in two-dimensional feature space with coordinate axes $x_1, x_2$. Regions of low data coverage exhibit an increased Mahalanobis distance with respect to $A_t^{-1}$ and therefore yield a stronger exploration bonus.}
  \label{fig:geom}
\end{figure}
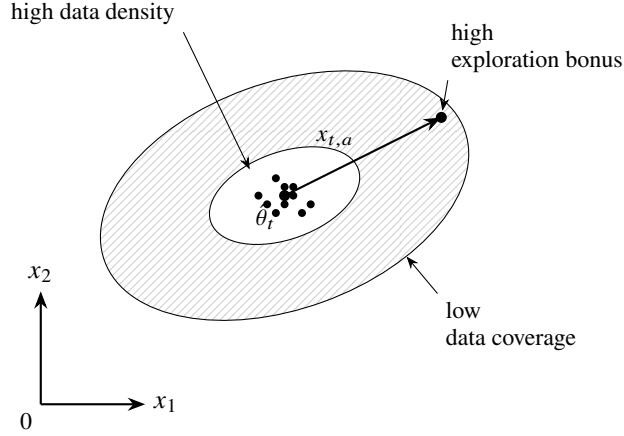

Exploration occurs non-isotropically along feature space directions where epistemic model uncertainty is maximal. Context vectors $x_{t,a}$ pointing toward regions of low data coverage exhibit increased Mahalanobis distance, yielding a higher exploration bonus to prioritize under-explored content and mitigate popularity bias.

\subsection{Numerical Stability via Direct Cholesky Updates}
LinUCB efficiency depends heavily on handling matrix $A_t$. Conventional implementations rely on the Sherman-Morrison formula for recursive matrix inversion. Under real-world conditions, this is numerically unstable: floating-point subtractions cause loss of significance and error accumulation, destroying the SPD property of the inverse.

Kairos avoids Sherman-Morrison inversions by executing rank-1 updates directly on triangular Cholesky factors. Since $A_t$ is a Symmetric Positive Definite (SPD) matrix, the factorization $A_t = L_t L_t^\top$ exists ($L_t \in \mathbb{R}^{d \times d}$ lower triangular). Upon receiving a new context vector $x \in \mathbb{R}^d$ and reward $r \in \mathbb{R}$, $L_t$ is updated via rank-1 modification \cite{Gi74}. This provides structural regularization, preventing negative eigenvalues and guaranteeing SPD properties implicitly.

Incremental updates are specified in \Cref{alg:cholesky} in detail.

\begin{algorithm}[H]
  \caption{Incremental Cholesky Ranking Update}
  \label{alg:cholesky}
  \begin{algorithmic}[1]
    \Procedure{UpdateModel}{$L_t, b_t, x, r, \gamma$}
    \State $x \gets x / \|x\|$ \Comment{normalize feature vector $x \in \mathbb{R}^d$}
    \State $L_t \gets \sqrt{\gamma} \cdot L_t, \quad b_t \gets \gamma \cdot b_t$ \Comment{apply discount factor $\gamma \in (0,1]$}
    \If{$r > 0$}
    \State $L_{t+1} \gets \text{cholupdate}(L_t, \sqrt{r} \cdot x)$ \Comment{rank-1 Cholesky update: $L_{t+1} L_{t+1}^\top = L_t L_t^\top + r x x^\top$}
    \EndIf
    \State $b_{t+1} \gets b_t + r \cdot x$ \Comment{update reward vector $b \in \mathbb{R}^d$}
    \State $y \gets L_{t+1} \backslash x$ \Comment{forward substitution for $y \in \mathbb{R}^d$}
    \State $\sigma^2 \gets y^\top y$ \Comment{uncertainty: $\sigma^2 = x^\top A_{t+1}^{-1} x$}
    \State \Return $L_{t+1}, b_{t+1}, \sigma^2$
    \EndProcedure
  \end{algorithmic}
\end{algorithm}

As shown in \Cref{alg:cholesky}, the variables represent the following mathematical objects:
\begin{itemize}
  \item $L_t \in \mathbb{R}^{d \times d}$: Lower triangular Cholesky factor of covariance matrix ($A_t = L_t L_t^\top$).
  \item $b_t \in \mathbb{R}^d$: Accumulated weighted context rewards ($b_t = \sum_{\tau=1}^{t-1} r_\tau x_{\tau}$).
  \item $x \in \mathbb{R}^d$: Normalized article context vector ($\|x\|_2 = 1$).
  \item $r \in \mathbb{R}$: Observed scalar reward signal ($r \in \{0, 1\}$).
  \item $\gamma \in (0,1]$: Discount factor for temporal decay of historical interactions.
  \item $y \in \mathbb{R}^d$: Forward substitution solution $L_{t+1} y = x$, directly yielding uncertainty $\sigma^2 = y^\top y = x^\top A_{t+1}^{-1} x$ without explicit inversion.
\end{itemize}

Updating in Cholesky factor space strictly guarantees SPD properties of $A_t$.

\subsection{Computational Effort and Complexity}
Let $d \in \mathbb{N}$ denote feature space dimension with $d = \metricDimMRL$ in the MRL subspace and $d = \metricDimFull$ in full space. Although both Sherman-Morrison inversion and Cholesky rank-1 update have an asymptotic time complexity of $\mathcal{O}(d^2)$ per iteration, practical robustness differs significantly.

While matrix inversion is vulnerable to poor condition numbers of $A_t$ (requiring expensive re-initializations), Cholesky updates maintain stability in triangular factor space. Memory footprint remains $\mathcal{O}(d^2)$ for $L_t$, but numerical hygiene enables continuous operation without manual intervention.

\subsection{Matryoshka Representation Learning}
To enable adaptive inference without redundant computations, Kairos employs Matryoshka Representation Learning (MRL) \cite{Ku22} via Nomic Embed \cite{Nu24}. The model learns embeddings $\Phi(x) \in \mathbb{R}^{\metricDimFull}$ exhibiting high representativeness across nested subspaces:
\begin{equation}
  m \in \mathcal{M} = \{64, \metricDimMRL, 256, 512, \metricDimFull\}
\end{equation}

Optimization uses a nested loss function:
\begin{equation}
  \mathcal{L}_{\text{MRL}}(x; \Phi) = \sum_{m \in \mathcal{M}} c_m \cdot \mathcal{L}\left(W^{(m)} \cdot \Phi(x)^{(m)}\right)
  \label{eq:mrl}
\end{equation}
where $\Phi(x)^{(m)} \in \mathbb{R}^m$ is the feature vector truncated to the first $m$ dimensions, $W^{(m)}$ is the classification head weight matrix, and $c_m > 0$ is subspace weight.

In Kairos, Maximum Inner Product Search (MIPS) candidate generation runs on an $m = \metricDimMRL$ subspace, significantly reducing compute complexity while preserving semantic depth for final ranking.

\section{Evaluation}
Empirical validation was performed under realistic data-scarcity conditions: an article corpus ($N = \metricDatasetSize$) reflecting regional news outlet volume within a \metricTTL{} window, representing an exemplary scenario for which Kairos was designed.

\subsection{Spectral Analysis and Information Preservation}
\begin{figure}[!htbp]
  \centering
  \includegraphics[width=0.62\linewidth]{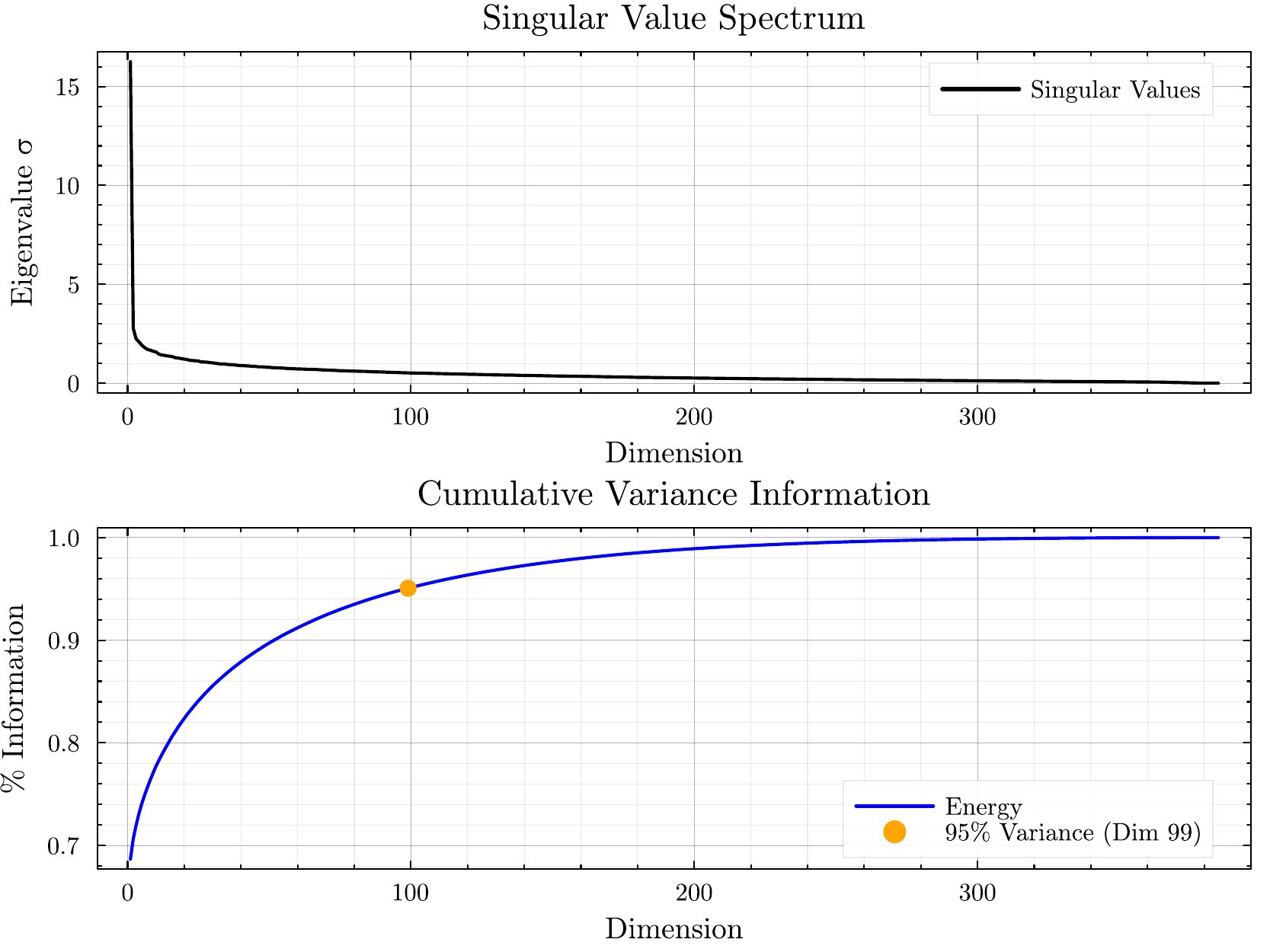}
  \caption{Spectral analysis of news corpus ($N=\metricDatasetSize$). Scree plot (top) displays rapid singular value decay; cumulative variance (bottom) shows ~\metricSVDVariance{} semantic energy concentrated in the first \metricSVDDim{} dimensions.}
  \label{fig:spektral}
\end{figure}

Spectral analysis (\Cref{fig:spektral}) confirms that at $m = \metricDimMRL$ dimensions, over \metricSVDVariance{} of semantic variance is preserved, indicating strong semantic redundancy in news feature spaces. Dimensionality reduction filters correlated noise while retaining retrieval signals.

\FloatBarrier

\subsection{Computational Efficiency and Approximation Quality}
Inference latency was benchmarked using Julia's \texttt{BenchmarkTools.jl}. To ensure statistical significance, auto-tuning accumulated measurements over 5s windows or up to 10,000 samples per configuration.

Approximation quality was evaluated via cosine similarity deviations across all \metricPairCount{} article pairs. The mean absolute error (MAE) of $\metricMAE$ confirms \metricPreservationPercent{} structural retention, demonstrating MRL efficiency in isolating signal from noise.

\begin{figure}[!htbp]
  \centering
  \includegraphics[width=0.56\linewidth]{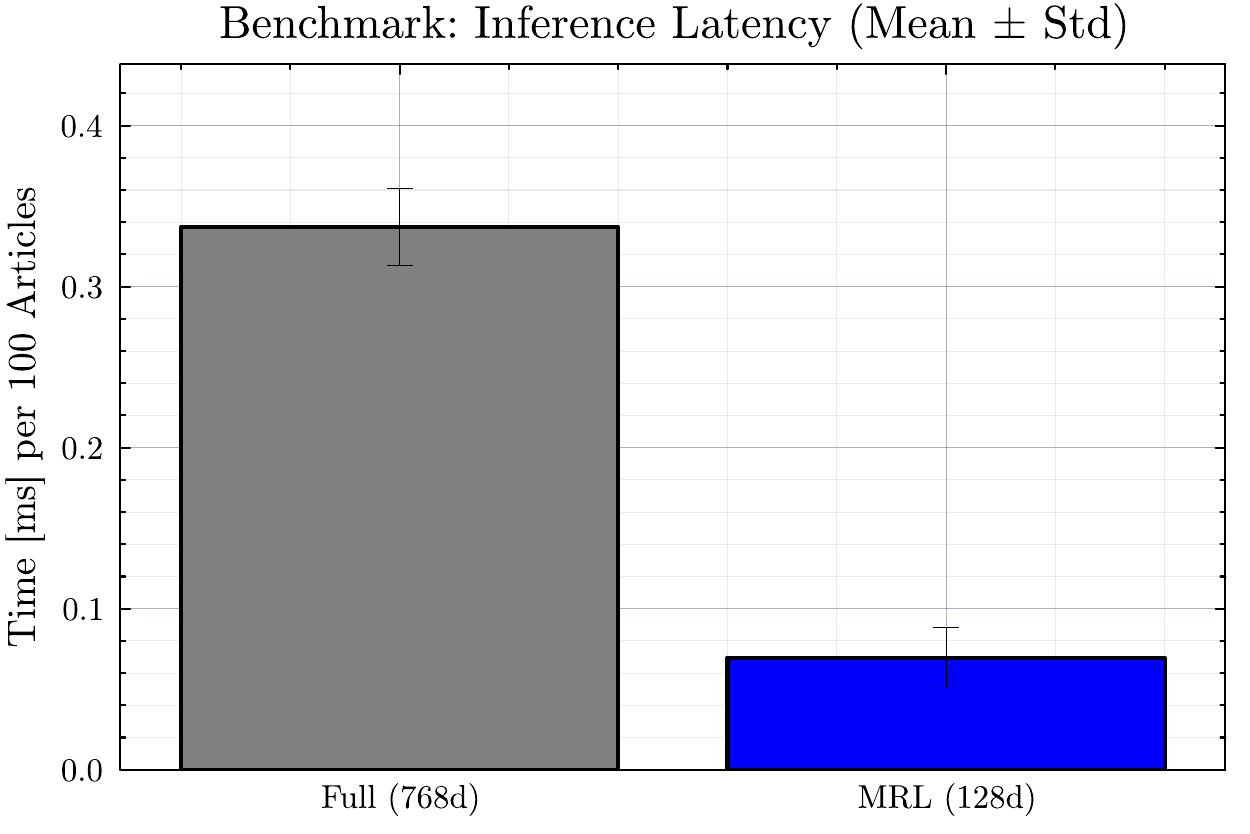}
  \caption{Inference latency benchmark. \metricDimMRL d subspace reduction achieves \metricSavingsPercent{} compute savings ($\metricSpeedupTimes$ speedup).}
  \label{fig:latency}
\end{figure}

For news streaming, top-$k$ candidates are retrieved with negligible error while boosting system inference capacity nearly 5-fold. Full \metricDimFull d space incurs $\metricLatencyFull$\,ms latency per 100 articles, whereas the MRL \metricDimMRL d subspace reduces latency to $\metricLatencyMRL$\,ms—a \metricSavingsOver{} compute reduction (\metricSpeedupFactor{} speedup, \Cref{fig:latency}).

\subsection{Reproducibility}
To ensure full reproducibility, source code implemented in Julia 1.10, Tagesschau API preprocessing scripts, configuration files, and random seeds are publicly available in the dedicated GitHub repository.

\section{Discussion}

\subsection{Numerical Resilience and Algorithmic Stability}
Cholesky-based implementation benefits manifest primarily in maintaining numerical integrity under real operational conditions. Regional news rewards are noisy, yielding ill-conditioned context matrices under sparse data.

While classical Sherman-Morrison updates degrade condition numbers through floating-point subtractions \cite{GV13}, Kairos updates directly in triangular factor space, avoiding error propagation and guaranteeing matrix positive definiteness \cite{Hi02}.

\begin{figure}[!htbp]
  \centering
  \includegraphics[width=0.64\linewidth]{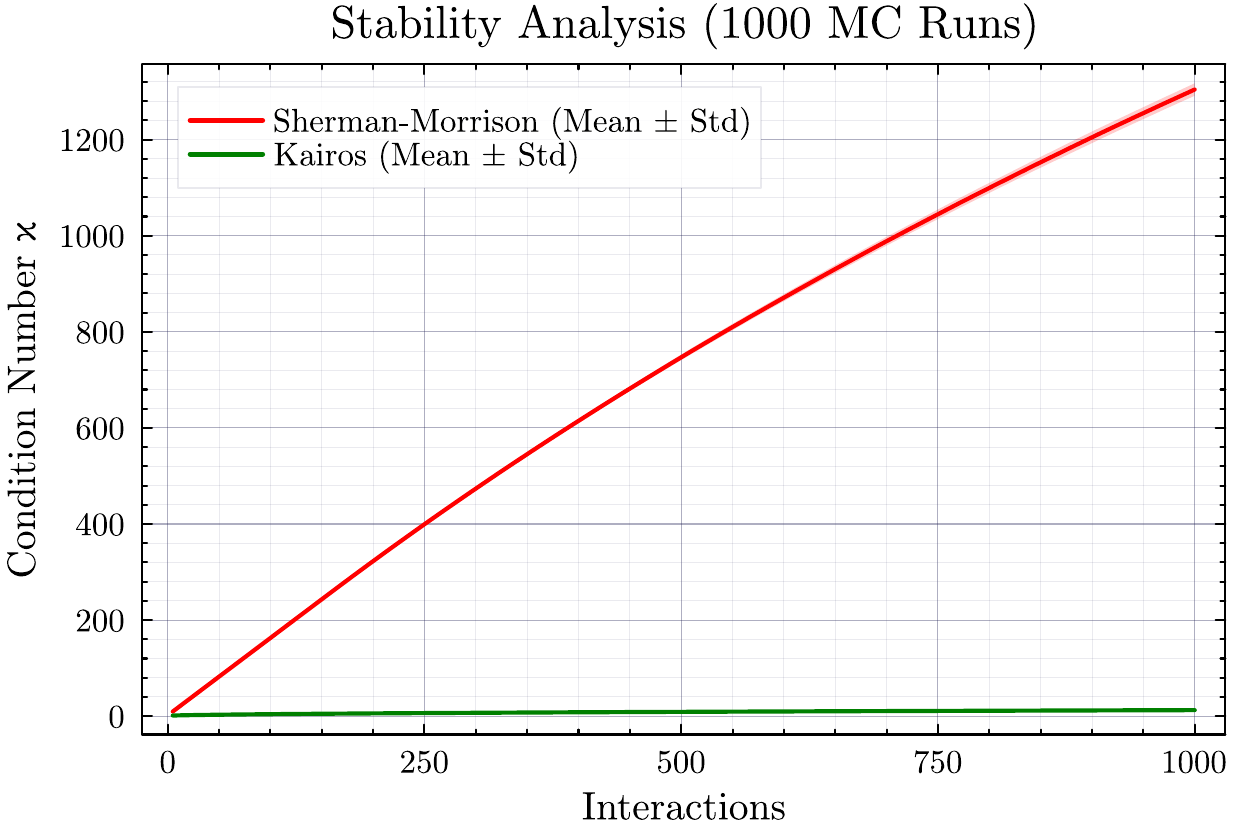}
  \caption{Comparative stability analysis over \metricMCRuns{} Monte Carlo simulations. Classical Sherman-Morrison inversion (red) exhibits condition number divergence and expanding variance, whereas the Cholesky approach (green) maintains near-zero error variance.}
  \label{fig:stability}
\end{figure}

Monte Carlo analysis over \metricMCRuns{} runs (\Cref{fig:stability}) confirms that Kairos maintains numerical consistency under high load, whereas conventional inversion diverges.

Additionally, contextual bandit exploration mitigates popularity bias, giving fresh articles early exposure during their short TTL.

\subsection{The Precision-Efficiency Trade-off}
The $\metricSpeedupTimes$ speedup with \metricMAEVal{} MAE shows full \metricDimFull d representations are redundant for retrieval. Asymmetric workload distribution (\metricDimMRL d for MIPS, full precision for LinUCB ranking) scales user capacity without proportional hardware costs.

\subsection{Perspective Extensions}
The Cholesky foundation enables geometric learning constraints, such as manifold regularization for Lipschitz bounds \cite{Be25}, bounding sensitivity to anomalous reward signals (e.g., click bots).

\subsection{Limitations of the Evaluation}
Evaluation relies on a small real-time corpus ($N=\metricDatasetSize$), limiting generalizability to large production platforms. User-centric quality metrics (CTR, long-term diversity) and live user interaction rewards remain subjects for future online testing.

\section{Conclusion and Outlook}
Project Kairos demonstrates a robust approach to overcoming the item cold-start paradox in news recommendation by combining numerically stable online learning with adaptive Matryoshka feature representations.

Key findings: Cholesky covariance updates mathematically secure LinUCB optimization under high load; MRL embeddings yield a $\metricSpeedupTimes$ speedup with minimal precision loss; contextual exploration maximizes utility during short article TTLs.

Future work will focus on formal stability proofs, manifold constraints \cite{Be25} for Lipschitz guarantees against adversarial signals, and GPU/hardware acceleration of Cholesky kernels for massive parallel real-time streams. Project Kairos forms a foundation for next-generation, mathematically secured recommendation systems.

\clearpage

\printbibliography

\end{document}